\documentclass{article}

\usepackage{arxiv}

\usepackage[utf8]{inputenc} 
\usepackage[T1]{fontenc}    
\usepackage{hyperref}       
\usepackage{url}            
\usepackage{booktabs}       
\usepackage{amsfonts}       
\usepackage{nicefrac}       
\usepackage{microtype}      
\usepackage{lipsum}
\usepackage{svg}
\usepackage{graphicx}
\graphicspath{ {./images/} }
\usepackage{amsmath} 
\usepackage{pdflscape}
\usepackage{xcolor}
\usepackage[normalem]{ulem}
\usepackage{comment}
 \usepackage{longtable}
\usepackage{tabularx}
\usepackage{caption} 
\usepackage{subcaption}
\usepackage{longtable}
\usepackage{booktabs}
\usepackage[utf8]{inputenc}

\title{What is YOLOv5: A deep look into the internal features of the popular object detector}

\author{
  \textbf{Rahima Khanam}\textsuperscript{*} and \textbf{Muhammad Hussain}\\[1ex] 
  \begin{minipage}[t]{0.90\textwidth}
    \scriptsize Department of Computer Science, Huddersfield University, Queensgate, Huddersfield HD1 3DH, UK; \\
    \textsuperscript{*}Correspondence: rahima.khanam@hud.ac.uk;
  \end{minipage}
}
\begin{document}
\maketitle
\begin{abstract}This study presents a comprehensive analysis of the YOLOv5 object detection model, examining its architecture, training methodologies, and performance. Key components, including the Cross Stage Partial backbone and Path Aggregation-Network, are explored in detail. The paper reviews the model's performance across various metrics and hardware platforms. Additionally, the study discusses the transition from Darknet to PyTorch and its impact on model development. Overall, this research provides insights into YOLOv5's capabilities and its position within the broader landscape of object detection and why it is a popular choice for constrained edge deployment scenarios.
\end{abstract}

\keywords{Automation; Computer Vision; YOLO; YOLOV5; Object Detection; Real-Time Image processing; Convolutional Neural Networks(CNN); YOLO version comparison}

\section{Introduction}
Computer vision is a rapidly evolving field that empowers machines~\cite{14} to interpret and comprehend visual information~\cite{hussain2024yolov1}. A critical component of this discipline is object detection~\cite{zahid2023lightweight}, which involves accurately identifying and locating objects within images or video sequences~\cite{hussain2023and}. To address this challenge, various algorithmic approaches have been developed, with significant advancements achieved in recent years\cite{hussain2024depth}.

One such groundbreaking technique is the You Only Look Once (YOLO) algorithm, introduced by Redmon et al. in 2015~\cite{redmon2016you}. This name reflects its distinctive approach, examining the entire image just once to identify objects and their positions. In contrast to conventional methods employing two-stage detection processes, YOLO treats object detection as a regression problem~\cite{redmon2016you}. In the YOLO paradigm, a single convolutional neural network is employed to predict bounding boxes and class probabilities for an entire image~\cite{hussain2023yolo1}. This streamlined approach differs from traditional methods with more intricate pipelines.

Building upon the foundation laid by YOLO, this paper delves into the YOLOv5 architecture, a state-of-the-art object detection model that has garnered significant attention due to its exceptional performance and efficiency.

\subsection{Survey Objective} 

This study aims to evaluate the performance of the YOLOv5 object detection model in comparison to state-of-the-art alternatives. The research focuses on assessing the trade-off between model accuracy and inference speed across the various YOLOv5 variants (n, s, m, l, x). By examining these different model sizes, this study seeks to identify the optimal balance between performance metrics for diverse application requirements.

The investigation will explore the factors contributing to YOLOv5's performance gains, with particular emphasis on:

\begin{enumerate}
    \item The role of PyTorch training methodologies
    \item The impact of architectural innovations such as the CSP backbone and PA-Net neck
    \item The effectiveness of data augmentation techniques, including mosaic augmentation
    \item The influence of loss calculation methods and bounding box anchor generation
\end{enumerate}

Additionally, the study will analyze the implications of YOLOv5's transition from the Darknet framework to PyTorch, considering aspects such as model development, deployment, and accessibility for custom object detection tasks.

By providing a comprehensive examination of YOLOv5's capabilities and innovations, this research aims to contribute to the broader understanding of advanced object detection algorithms and their practical applications in computer vision.

\section{Evolution of Yolov5}
The YOLOv5 \cite{yolov5Github} repository emerged as an evolution of the YOLOv3 \cite{yolov3Github} PyTorch implementation developed by Glenn Jocher in 2020. It was released following the release of YOLOv4~\cite{Yao2021}. Yolov5 gained popularity as a platform for transitioning YOLOv3 models from Darknet to PyTorch for production deployment. Recognizing the utility of this PyTorch-based approach, Ultralytics initiated a development trajectory focused on enhancing the YOLOv3 architecture and training methodologies within the PyTorch framework. This endeavor aimed to empower a broader community of developers to create and deploy custom object detectors.

This is YOLOv5 development timeline:

\begin{itemize}
    \item \textbf{April 1, 2020:} Initiation of development on a series of compound-scaled PyTorch models based on YOLOv3/YOLOv4 architectures.
    \item \textbf{May 27, 2020:} Public release of the YOLOv5 repository, demonstrating state-of-the-art performance among existing YOLO implementations.
    \item \textbf{June 9, 2020:} Integration of CSP (Cross-Stage Partial) modules, contributing to improved model speed, size, and accuracy.
    \item \textbf{June 19, 2020:} Adoption of FP16 precision as default, resulting in smaller checkpoints and faster inference.
    \item \textbf{June 22, 2020:} Implementation of PANet updates, including new detection heads, reduced parameters, and enhanced mAP (mean Average Precision).
\end{itemize}

Initially, the model incorporating these advancements was designated YOLOv4, aligning with the contemporaneous release of the YOLOv4 architecture within the Darknet framework. However, to avoid potential confusion and inconsistencies, it was subsequently revised to YOLOv5. While this change sparked some debate within the research community, a comparative analysis of YOLOv4 and YOLOv5 was conducted to facilitate objective evaluation \cite{yolov4vsyolov5}.

This study primarily focuses on the novel techniques and performance metrics introduced within the YOLOv5 framework, as detailed within the GitHub repository. It is essential to acknowledge the rapid evolution of the YOLOv5 model since its inception, which suggests ongoing research and development in this domain. As such, YOLOv5 should be considered a dynamic and evolving system rather than a static model.

\section{Architectural footprint of Yolov5}
Object detection, a primary application of YOLOv5, entails the extraction of salient features from input images. These features are subsequently processed by a predictive model to localize and classify objects within the image.

The YOLO architecture introduced the end-to-end, differentiable approach to object detection by unifying the tasks of bounding box regression and object classification into a single neural network \cite{khanam2024comprehensive}. Fundamentally, the YOLO network comprises three core components. The \textbf{backbone}, a convolutional neural network, is responsible for encoding image information into feature maps at varying scales. These feature maps are then processed by the \textbf{neck}, a series of layers designed to integrate and refine feature representations. Finally, the \textbf{head} module generates predictions for object bounding boxes and class labels based on the processed features.

\begin{figure}[h]
    \centering
    \includegraphics[width=1\linewidth]{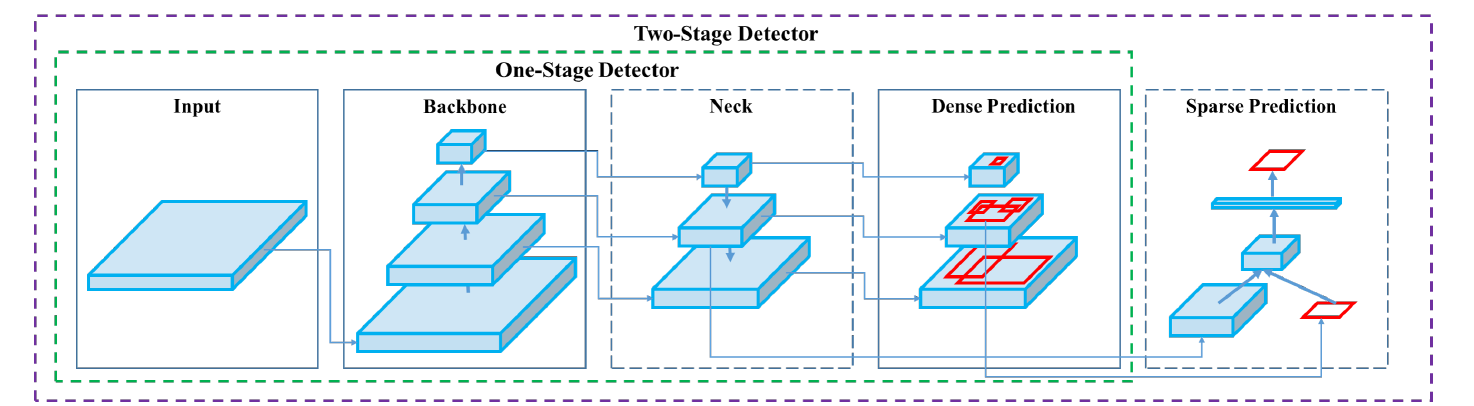}
    \caption{Process of Object Detection \cite{bochkovskiy2020yolov4}}
    \label{fig:ODprocess}
\end{figure}

While there exists considerable flexibility in the architectural design of each component, YOLOv4 and YOLOv5 have significantly advanced the field by incorporating techniques from other computer vision domains. This synergistic approach has demonstrated substantial improvements in object detection performance within the YOLO framework.

\subsection{YOLOv5 Training Methods}
The efficacy of an object detection system is contingent not only on its underlying architecture but also on the employed training methodologies. While architectural innovations often garner significant attention, the role of training procedures in achieving optimal performance is equally critical. There are two primary training techniques employed in YOLOv5:

\begin{itemize}
    \item \textbf{Data augmentation} is a pivotal component of the YOLOv5 training pipeline. By introducing diverse transformations to the training dataset, this technique enhances model robustness and generalization capabilities. Consequently, the model becomes more resilient to variations in real-world image conditions.
    \item The \textbf{Loss function} is a composite metric calculated from three primary components: Generalized Intersection over Union (GIoU), objectness, and classification loss. These loss components are carefully designed to optimize mean average precision (mAP), a widely adopted evaluation metric for object detection models.
\end{itemize}

\subsection{Transition to PyTorch}
YOLOv5 represents a significant advancement by transitioning the YOLO architecture from the Darknet framework to PyTorch \cite{pytorchBrandGuidelines}. The Darknet framework, primarily implemented in C, affords researchers granular control over network operations. While this level of control is advantageous for experimentation, it often hinders the rapid integration of novel research findings due to the necessity of custom gradient calculations for each new implementation.

Porting the training procedures of Darknet to PyTorch, as achieved in YOLOv3, is a complex undertaking in itself. YOLOv5 extends this effort by further optimizing and refining these procedures within the PyTorch ecosystem.

\subsection{Data augumentation}
YOLOv5 incorporates data augmentation techniques within its training pipeline to enhance model robustness and generalization. During each training epoch, images are subjected to a series of augmentations through an online data loader, including:

\begin{itemize}
    \item \textbf{Scaling:} Adjustments to image size.
    \item \textbf{Color space manipulation:} Modifications to color channels.
    \item \textbf{Mosaic augmentation:} A novel technique that combines four images into four randomly sized tiles.
\end{itemize}.

Introduced in the YOLOv3 PyTorch repository and subsequently integrated into YOLOv5, mosaic data augmentation has proven particularly effective in addressing the challenge of small object detection, a common limitation in datasets by the Common Objects in Context (COCO) object detection benchmark. By combining multiple images into a single training example, this technique exposes the model to a wider variety of object scales and spatial arrangements, thereby enhancing its ability to accurately detect smaller objects.

\subsection{Bounding Box Anchors}
The YOLOv3 PyTorch repository introduced a novel approach to anchor box generation, employing K-means clustering and genetic algorithms to derive anchor box dimensions directly from the distribution of bounding boxes within a given dataset. This methodology is particularly critical for custom object detection tasks, as the scale and aspect ratios of objects often diverge significantly from those commonly found in standard datasets like COCO.

The YOLOv5 architecture predicts bounding box coordinates as offsets relative to a predefined set of anchor box dimensions. These anchor dimensions are essential for initializing the prediction process and can significantly influence the model's performance.

\begin{figure}[h]
    \centering
    \includegraphics[width=0.5\linewidth]{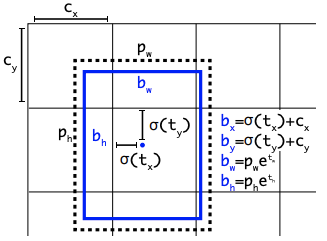}
    \caption{Bounding box prediction based on an anchor box \cite{anchorboxRoboflow}}
    \label{fig:enter-label}
\end{figure}

\subsection{Loss Calculation}
The YOLOv5 loss function is a composite of three components: Binary Cross-Entropy (BCE) for class prediction and objectness, and Complete Intersection over Union (CIoU) for localization. The overall loss is computed as a weighted sum of these individual losses:

\begin{equation}
\text{Loss} = \lambda_1 \cdot L_{\text{cls}} + \lambda_2 \cdot L_{\text{obj}} + \lambda_3 \cdot L_{\text{loc}}
\end{equation}

where $L_{\text{cls}}$, $L_{\text{obj}}$, and $L_{\text{loc}}$ represent the BCE loss for class prediction, BCE loss for objectness, and CIoU loss for localization, respectively. The coefficients $\lambda_1$, $\lambda_2$, and $\lambda_3$ are hyperparameters that balance the contributions of each loss component to the overall optimization process.

\subsection{16 Bit Floating Point Precision}
The PyTorch framework offers the capability to reduce computational precision from 32-bit to 16-bit floating-point numbers during both training and inference. When applied to YOLOv5, this technique has demonstrated potential for significant acceleration in inference speed. However, it is essential to note that these performance gains is only contingent upon specific GPU architectures, notably the V100 and T4 models for Yolov5. While the limitations of hardware compatibility exist at present, ongoing developments by NVIDIA indicate a prospective expansion of this efficiency enhancement to a wider range of hardware platforms.

\subsection{CSP Backbone}
Both YOLOv4 and YOLOv5 incorporate the CSP (Cross-Stage Partial) bottleneck module for feature extraction. This architectural innovation, introduced by WongKinYiu et al., addresses the issue of redundant gradient information prevalent in larger convolutional neural network backbones. By decoupling feature maps into two main parts and recombining them, the CSP module effectively reduces computational cost and model complexity without compromising performance. This efficiency enhancement is particularly advantageous for the YOLO family, where rapid inference and compact model size are paramount.

\begin{figure}[h]
    \centering
    \includegraphics[width=0.5\linewidth]{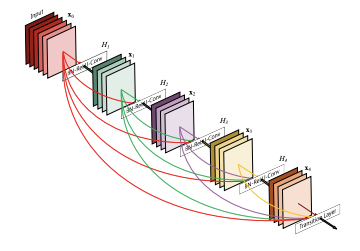}
    \caption{Interconnected dense layers in DenseNet \cite{huang2017densely}}
    \label{fig:denseNet}
\end{figure}

CSP models draw inspiration from the DenseNet architecture as shown in Figure \ref{fig:denseNet}, addressing the challenges inherent in deep CNNs, including the vanishing gradient problem. By fostering direct connections between layers, DenseNet aimed to enhance feature propagation, promote feature reuse, and reduce the overall number of network parameters.

\begin{figure}[h]
    \centering
    \includegraphics[width=1\linewidth]{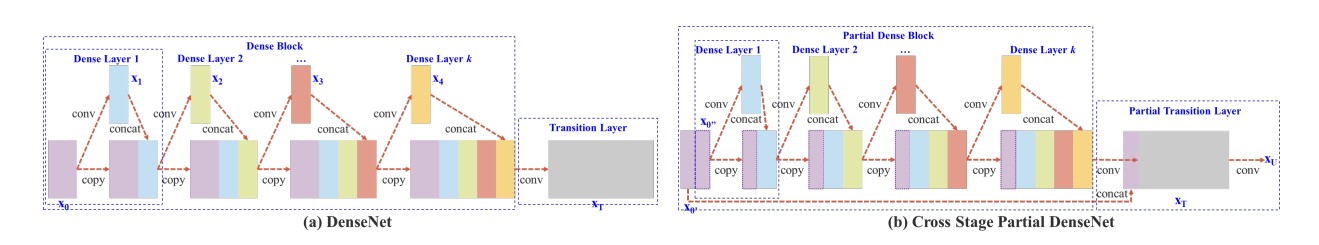}
    \caption{(a) DenseNet and (b) CSPDenseNet \cite{wang2020cspnet}}
    \label{fig:CSPDenseNet}
\end{figure}

The CSPResNext50 and CSPDarknet53 architectures incorporate modifications to the DenseNet structure as illustrated in Figure \ref{fig:CSPDenseNet}. Specifically, the feature maps generated by the base layer are divided into two branches. One branch is processed through a dense block, while the other is forwarded directly to the subsequent stage. This architectural modification aims to mitigate computational bottlenecks inherent to DenseNet and enhance learning by preserving an unaltered representation of the original feature map.

\subsection{PA-Net Neck}
\begin{figure}[h]
    \centering
    \includegraphics[width=1\linewidth]{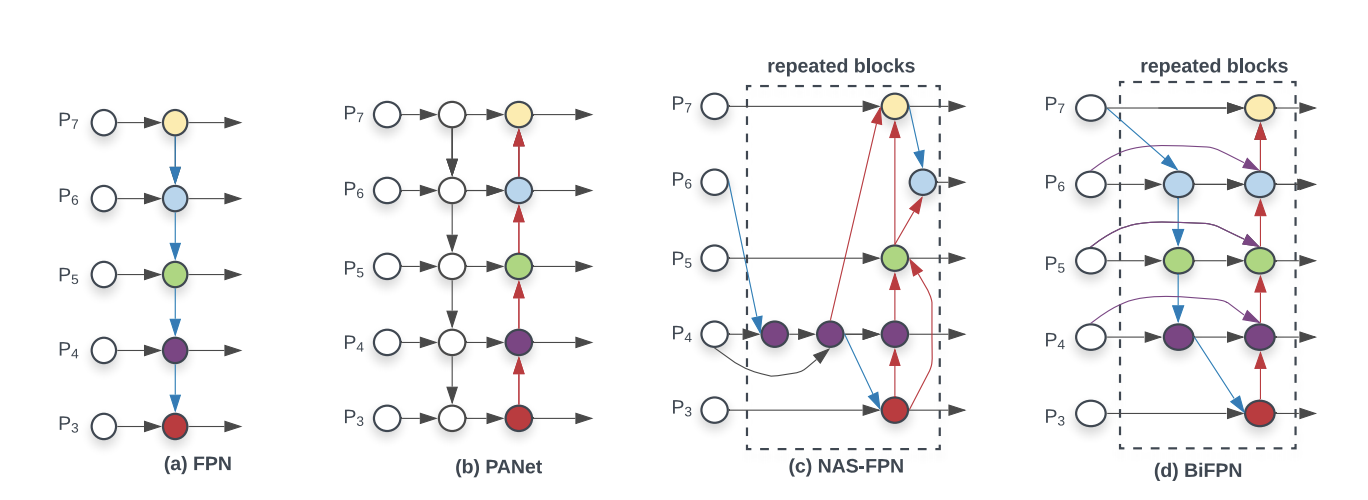}
    \caption{Variations of FPN architectures \cite{tan2020efficientdet}}
    \label{fig:PANNET}
\end{figure}

Both YOLOv4 and YOLOv5 employ the PA-Net architecture for feature aggregation \cite{Solawetz2020}. As illustrated in Figure \ref{fig:PANNET}, each P\_i represents a distinct feature layer extracted from the CSP backbone. This architectural choice is inspired by the EfficientDet object detection framework, where the BiFPN module was deemed optimal for feature integration. While the BiFPN approach serves as a benchmark, it is plausible that alternative implementations within the YOLO framework could yield further performance improvements.

It is evident that YOLOv5 builds upon the foundational research of YOLOv4 in selecting the most suitable neck architecture. YOLOv4 comprehensively evaluated a range of options including FPN, PAN, NAS-FPN, BiFPN, ASFF, and SFAM.

\section{YOLOv5 Models}
The YOLOv5 architecture encompasses five distinct models, ranging from the computationally efficient YOLOv5n to the high-precision YOLOv5x.

\begin{itemize}
    \item \textbf{YOLOv5n:} Designed for resource-constrained environments, YOLOv5n is the smallest and fastest model in the series. Its compact size, less than 2.5 MB in INT8 format and approximately 4 MB in FP32 format, makes it suitable for deployment on edge devices and IoT platforms. Compatibility with OpenCV DNN further enhances its utility for mobile applications.
    \item \textbf{YOLOv5s:} Representing the baseline model, YOLOv5s incorporates approximately 7.2 million parameters. Its performance characteristics make it suitable for CPU-based inference tasks.
    \item \textbf{YOLOv5m:} Positioned as a mid-sized model with 21.2 million parameters, YOLOv5m offers a balanced combination of speed and accuracy. This model is often considered a versatile choice for a wide range of object detection applications and datasets.
    \item \textbf{YOLOv5l:} With 46.5 million parameters, YOLOv5l is designed for scenarios requiring higher precision, particularly in detecting smaller objects within images.
    \item \textbf{YOLOv5x:} As the largest and most complex model in the series, YOLOv5x achieves the highest mAP among its counterparts. However, this performance comes at the cost of increased computational requirements due to its 86.7 million parameters.
\end{itemize}

A comprehensive overview of the model variants, encompassing inference speed across CPU and GPU platforms as well as the number of parameters for a 640-pixel image size, is presented in Table \ref{tab:modelvariants}.

\begin{table}[h]
\centering
\caption{YOLOv5 Model Performance}
\label{tab:modelvariants}
\begin{tabular}{|c|c|c|c|c|}
\hline
Model & Params (Million) & Accuracy (mAP 0.5) & CPU Time (ms) & GPU Time (ms) \\
\hline
YOLOv5n & 1.9 & 45.7 & 45 & 6.3 \\
YOLOv5s & 7.2 & 56.8 & 98 & 6.4 \\
YOLOv5m & 21.2 & 64.1 & 224 & 8.2 \\
YOLOv5l & 46.5 & 67.3 & 430 & 10.1 \\
YOLOv5x & 86.7 & 68.9 & 766 & 12.1 \\
\hline
\end{tabular}
\end{table}


\section{YOLOv5 Annotation Format}
YOLOv5 employs a PyTorch TXT annotation format that closely resembles the YOLO Darknet TXT standard, with the addition of a YAML file specifying model configuration and class labels. To facilitate compatibility with YOLOv5, annotation data generated from various tools may require conversion. Platforms such as Roboflow offer support for multiple annotation formats and can export data directly into the YOLOv5-compatible format. Popular annotation tools including VOTT, LabelImg, and CVAT can also be utilized, with appropriate data conversion steps.

\section{YOLOv5 Labelling tools}
For efficient data management and annotation, Ultralytics, the developers of YOLOv5, recommends Roboflow as a compatible labeling tool \cite{yolov5partnership}. Table \ref{tab:yolov5_integrations} provides an overview of third-party integration platforms compatible with the YOLOv5. For each platform, the table outlines its primary functionalities when integrated with YOLOv5.

\begin{table}[h]
\centering
\caption{YOLOv5 Integrations}
\label{tab:yolov5_integrations}
\begin{tabular}{|c|l|}
\hline
Integration Platform & Functionality \\
\hline
Deci \cite{deci2024} & Automated compilation and quantization of YOLOv5 for enhanced inference performance. \\
ClearML \cite{clearml2024} & Tracking, visualization, and remote training of YOLOv5 models using an open-source platform. \\
Roboflow \cite{Solawetz2020} & Labeling and exporting custom datasets directly compatible with YOLOv5 training. \\
Weights and Biases \cite{wandb_yolov5} & Tracking and visualization of YOLOv5 training runs in the cloud. \\
\hline
\end{tabular}
\end{table}

\section{Discussion}
YOLOv5 represents a significant advancement in the field of object detection, building upon the foundations laid by its predecessors while introducing novel improvements are listed below:
\begin{enumerate}
    \item \textbf{Architectural Advancements:} YOLOv5 represents a significant advancement in object detection, building upon previous YOLO iterations. The incorporation of the CSP backbone and PA-Net neck enhances computational efficiency without compromising accuracy. The CSP module effectively addresses redundant gradient information, while the PA-Net architecture optimizes feature aggregation.
    \item \textbf{Model Versatility:} The range of YOLOv5 models (n, s, m, l, x) provides flexibility for diverse hardware and application requirements. The smallest model, YOLOv5n, expands object detection capabilities to edge devices and IoT platforms.
    \item \textbf{Training Methodology Innovations:} YOLOv5's emphasis on data augmentation, particularly mosaic augmentation, improves small object detection and reduces dataset size requirements. The adoption of 16-bit floating-point precision demonstrates a forward-looking approach to optimization.
    \item \textbf{Performance and Impact: }YOLOv5's high mAP scores and low inference times position it as a strong contender for real-time object detection. The transition to PyTorch has democratized access to the model, fostering broader research and development.
\end{enumerate}


\section{Conclusion}
YOLOv5 has emerged as a significant advancement in object detection, demonstrating a compelling balance of speed, accuracy, and user-friendliness. While the core architecture builds upon established principles, the model's implementation within the PyTorch framework represents a substantial leap forward, enhancing both development efficiency and deployment capabilities. The availability of multiple model variants tailored to diverse computational constraints expands YOLOv5's applicability across various domains from renewable energy~\cite{hussain2019deployment} to quality inspection in Manufacturing~\cite{hussain2023review}.

The model's accessibility, coupled with its compatibility with existing tools and platforms, positions YOLOv5 as a versatile solution for both research and practical applications. As the field of computer vision continues to evolve rapidly, YOLOv5 serves as a strong foundation for future advancements in real-time object detection and related tasks.

\vspace{6pt} 






\bibliographystyle{unsrt}  
\bibliography{ref}  

\end{document}